\documentclass[10pt,twocolumn,letterpaper]{article}

\usepackage{iccv}
\usepackage{times}
\usepackage{epsfig}
\usepackage{graphicx}
\usepackage{amsmath}
\usepackage{amssymb}

\usepackage[pagebackref=true,breaklinks=true,colorlinks,bookmarks=true]{hyperref} 
\usepackage[misc]{ifsym}

\iccvfinalcopy

\ificcvfinal\fi

\begin{document}

\title{Multi-Scale Memory Comparison for Zero-/Few-Shot Anomaly Detection}

\author{Chaoqin Huang$^{1,2,3}$\thanks{Chaoqin Huang and Aofan Jiang contributed equally.} \quad \quad Aofan Jiang$^{1,3*}$ \quad \quad Ya Zhang$^{1,3}$ \quad \quad Yanfeng Wang$^{1,3}$\thanks{Corresponding author.}\\ [2pt]
$^1$Cooperative Medianet Innovation Center, Shanghai Jiao Tong University, China\\
$^2$National University of Singapore, Singapore\\
$^3$Shanghai AI Laboratory, China\\
{\tt\small \{huangchaoqin,stillunnamed,ya\_zhang,wangyanfeng622\}@sjtu.edu.cn}}

\maketitle
\ificcvfinal\thispagestyle{empty}\fi

\begin{abstract}
   Anomaly detection has gained considerable attention due to its broad range of applications, particularly in industrial defect detection. To address the challenges of data collection, researchers have introduced zero-/few-shot anomaly detection techniques that require minimal normal images for each category. However, complex industrial scenarios often involve multiple objects, presenting a significant challenge. In light of this, we propose a straightforward yet powerful multi-scale memory comparison framework for zero-/few-shot anomaly detection. Our approach employs a global memory bank to capture features across the entire image, while an individual memory bank focuses on simplified scenes containing a single object. The efficacy of our method is validated by its remarkable achievement of 4th place in the zero-shot track and 2nd place in the few-shot track of the Visual Anomaly and Novelty Detection (VAND) competition.
\end{abstract}

\section{Introduction}

Visual anomaly detection and localization play a vital role in industrial manufacturing~\cite{huang2022ssm,ARNet}, aiming to classify and pinpoint defects using solely normal data. Within this context, zero-/few-normal-shot anomaly detection has gained significant importance in industrial settings. This approach offers promising advantages, including the potential to reduce data labeling and model training costs, as well as facilitating real-time anomaly detection. 

Early approaches to few-shot anomaly detection have explored different strategies such as data augmentation through transformations~\cite{TDG} or employing lightweight models to estimate normal distribution~\cite{DiffNet} in order to mitigate overfitting. Another notable technique, RegAD~\cite{regad}, introduced model reusing by pre-training an object-agnostic registration network on diverse images to capture normality for unseen objects. In more recent advancements, WinCLIP~\cite{winclip} has emerged as a notable method for zero-/few-shot anomaly detection. WinCLIP leverages the power of large-scale pre-trained open-source vision-language models such as CLIP~\cite{CLIP} to detect anomalies through prompt engineering. By capitalizing on the generalization ability of CLIP, WinCLIP demonstrates competitive performance in low-shot scenarios for both seen and unseen objects, approaching the performance levels of fully supervised methods. This extension of anomaly detection to zero-shot scenarios represents a significant advancement in the field. While WinCLIP exhibits impressive performance in zero-shot and few-shot anomaly detection tasks, its efficacy diminishes in complex scenes that involve multiple objects. The unconstrained feature extraction employed by CLIP does not guarantee the inclusion of information specific to unusual objects, particularly when only a single object within an image is anomalous.

\begin{table}[t]
    \centering
    \caption{Results of anomaly detection (F1-cls) and anomaly localization (F1-seg) on VAND challenge under the zero-shot and few-shot tracks, compared with state-of-the-art method WinCLIP.}
    \begin{tabular}{c|cccc}
    \hline
    & \multicolumn{2}{c}{Zero-Shot} & \multicolumn{2}{c}{Few-Shot}\\
         &  F1-cls & F1-seg&  F1-cls & F1-seg\\
         \hline
        WinCLIP & 0.7743 & 0.0953 & 0.8114 & 0.4118\\
        Ours & \textbf{0.7945} & \textbf{0.1866} & \textbf{0.8480} & \textbf{0.4515}\\
    \hline
    \end{tabular}
    \label{tab:vand}
\end{table}

To address these challenges, this paper presents a straightforward yet powerful multi-scale memory comparison framework for zero-/few-shot anomaly detection. Our approach incorporates a global memory bank to capture the features of the entire image in complex scenes, while an individual memory bank captures features in simplified scenes containing only a single object. To achieve this, we integrate the Segment Anything Model (SAM)~\cite{SAM} to determine the position of each individual object within an image. This segmentation process enables us to divide the multi-object anomaly detection problem into more manageable single-object anomaly detection tasks. 

For few-shot anomaly detection, we establish distinct multi-scale memory banks based on known normal images for each category. During inference, we compare the individual object instances with their corresponding individual memory bank, while the features of the entire image containing multiple objects are compared with the global memory bank. This separation of memory banks encourages the model to focus on specific challenges associated with single objects or multiple objects, resulting in enhanced anomaly detection performance. By combining the results obtained for multiple objects, we generate a comprehensive anomaly location map, where samples are considered anomalies when their features are not similar to those in the corresponding memory bank. Interestingly, we observe that even without the need for text prompts, our proposed method achieves comparable results on the segmentation task. This demonstrates the effectiveness of the multi-scale memory bank, significantly reducing the reliance on text prompt engineering, which typically requires substantial human resources. Under zero-shot conditions, where no normal images are provided, we leverage the language model to identify anomalies in both single-object and multi-object scenarios. Remarkably, our approach outperforms existing methods, showcasing superior results. This highlights the strength of our proposed framework in zero-shot anomaly detection, eliminating the need for extensive human labeling efforts.

We summarize the contributions as follows:
\begin{itemize}
    \item We present a multi-scale memory comparison framework specifically designed to overcome the limitations of existing zero-/few-shot anomaly detection methods in complex industrial defect detection scenes.
    \item Through comprehensive experiments, we demonstrate the effectiveness of our approach. Remarkably, our method achieved an impressive 4th place in the zero-shot track and a remarkable 2nd place in the few-shot track of the VAND (visual anomaly and novelty detection) competition. These results not only validate the competitive performance of our framework but also highlight its potential for practical applications in real-world scenarios.
\end{itemize}

\section{Method}
\subsection{Problem Statement}
We focus on unsupervised anomaly detection and localization in industrial visual inspection scenarios, where we have either no or very few normal samples available for reference. In this context, we are given a test image, denoted as $x$, and potentially a support set comprising $k$ normal samples, represented as $\{x_i\vert i = 1, ..., k\}$, all belonging to the same category. The primary objective is to predict whether the test image is anomalous or not, treating it as a binary classification task at the image level. Additionally, we aim to identify and segment the abnormal regions at the pixel level, providing a more fine-grained analysis of the anomalies present in the image.

\subsection{Method Overview}
Our proposed method leverages the strengths of two pretrained models, Contrastive Language Image Pre-training (CLIP) \cite{CLIP} and Segment Anything Model (SAM) \cite{SAM}, to create a powerful anomaly detection and segmentation system. Without the need for fine-tuning, our approach incorporates image and text prompt inputs and constructs multi-scale memories, enhancing the performance of the overall system.

\subsection{Zero-Shot Anomaly Detection}
\begin{figure*}[t]
    \centering
    \includegraphics[width=0.9\textwidth]{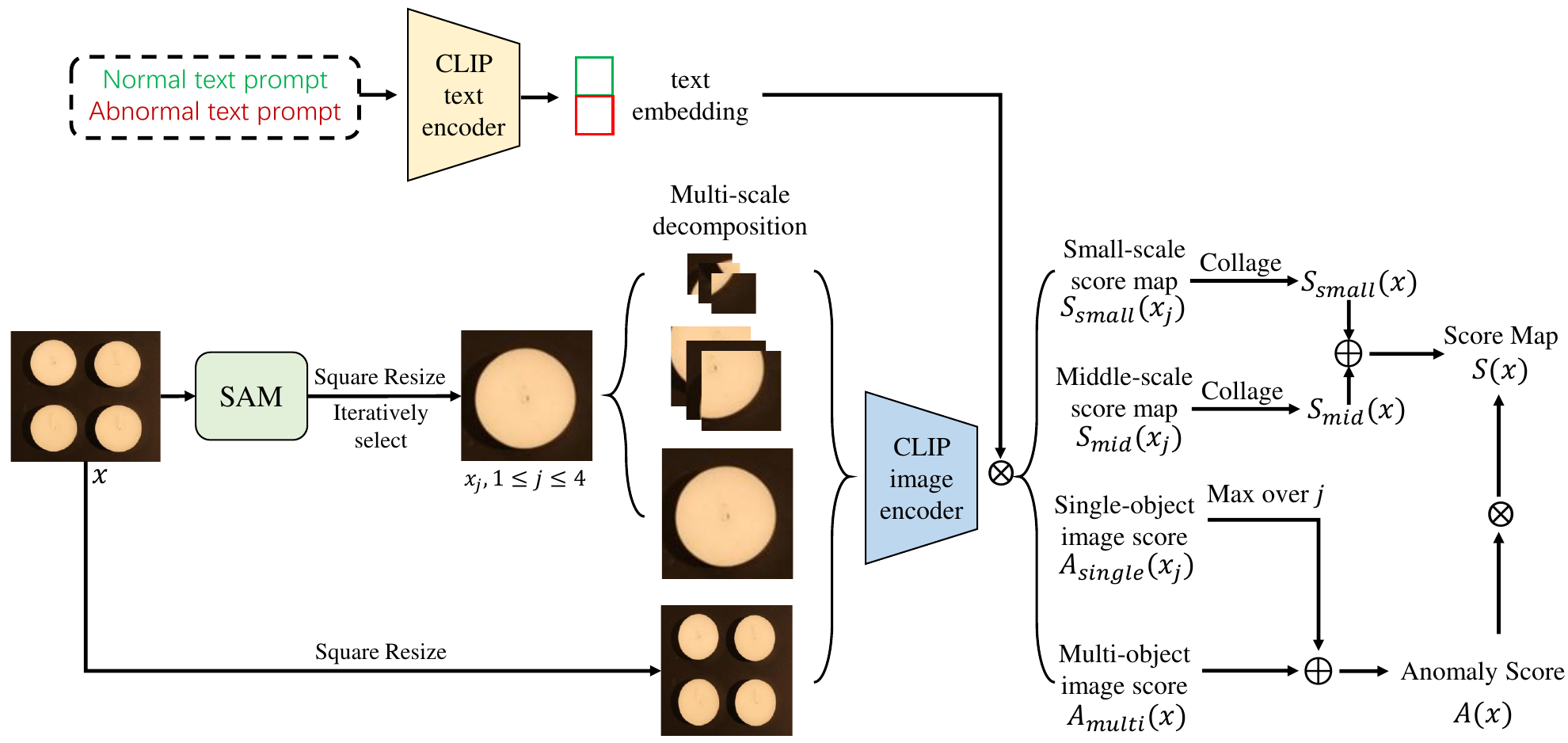}
    \caption{The overview of the proposed zero-shot anomaly detection and localization architecture.}
    \label{fig:zero}
\end{figure*}

In Figure \ref{fig:zero}, we illustrate our CLIP-based framework for zero-shot anomaly detection. The framework consists of two branches: the text branch and the image branch, which collaborate to detect anomalies. For the text branch, we utilize the method described in~\cite{winclip} to generate grouped text embeddings. These embeddings effectively capture the semantic details of the input text and serve as a reference for comparison.

\textbf{Individual Object Decomposition.} In the image branch, we proceed by segmenting the given test image $x$ into a set of individual square-resized images, each containing a single object. This segmentation is achieved using the Segment Anything Model (SAM), resulting in segmented images denoted as $\{x_j\vert j = 1, ..., n, n=\# \text{ of obj.}\}$.

\textbf{Visual Feature Extraction:} To effectively utilize individual images, we select each image $x_j$ in an iterative manner and perform multi-scale decomposition. This decomposition process employs a sliding-window strategy, whereby the image is masked separately using small-scale (2 × 2 patches), middle-scale (3 × 3 patches), and image-scale (same size as the individual image) windows with overlap. These patch scales (16 × 16 pixels) align with the patch scales used in the Vision Transformer (ViT). Following this decomposition process, we obtain a set of images for each individual image, with each image being masked at one consecutive region. Before forwarding these images, we discard the unmasked patches, akin to a masked autoencoder. This enables us to leverage the output class tokens to represent the features of the masked region. These tokens, combined with the original image class token, are then compared to the text embedding through matrix multiplication. The resulting similarity to the anomaly text embedding serves as the anomaly score for the corresponding input.

\textbf{Anomaly Detection:} To perform anomaly detection, we aggregate scores from both single object images and multiple object images to obtain the final anomaly score $A(x)$. This score is computed by adding the anomaly score from the multiple object image ($A_{multi}(x)$) and the maximum anomaly score from the individual object images ($A_{single}(x_j)$), where $j$ ranges from 1 to the total number of objects. Mathematically, this calculation can be represented as follows:
\begin{equation}
    A(x) = A_{multi}(x) + \max_j A_{single}(x_j).
\end{equation}
This ensemble approach enhances the overall capability of our framework for anomaly detection.

\textbf{Anomaly Segmentation:} For the anomaly segmentation task, we assign scores to all pixels within each masked window. In cases where windows overlap, the value within the overlap area is calculated using harmonic averaging. Consider an image $x_j$ with $t$ middle windows overlapped at pixel $i$. The middle-scale score for this pixel $S_{mid}(x_j)_i$ can be mathematically expressed as:
\begin{equation}
S_{mid}(x_j)_i = \frac{t}{\sum_t \frac{1}{score_t}}.
\end{equation}
. Following this approach, we obtain small-scale score map ($S_{small}(x_j)$) and middle-scale ($S_{mid}(x_j)$) score map for each individual image. To generate complete multi-scale score maps, we overlay these individual score maps onto their respective positions in the original image. These multi-scale score maps are then combined to form the final score map $S(x)$, which represents the overall likelihood of anomalies throughout the entire image. Specifically, the final score map $S(x)$ is computed as the sum of the small-scale score map $S_{small}(x)$ and the middle-scale score map $S_{mid}(x)$, weighted by anomaly score $A(x)$, \emph{i.e.},
\begin{equation}
    S(x) = A(x)\cdot (\lambda_1 S_{small}(x) + \lambda_2 S_{mid}(x)),
\end{equation}
where $\lambda_1$ and $\lambda_2$ are trade-off parameters weighting the multi-scale score maps, we adopt $\lambda_1 = 1.8, \lambda_2 = 0.2$ as the default.

\subsection{Few-Shot Anomaly Detection}

\begin{figure*}[t]
    \centering
    \includegraphics[width=1.0\textwidth]{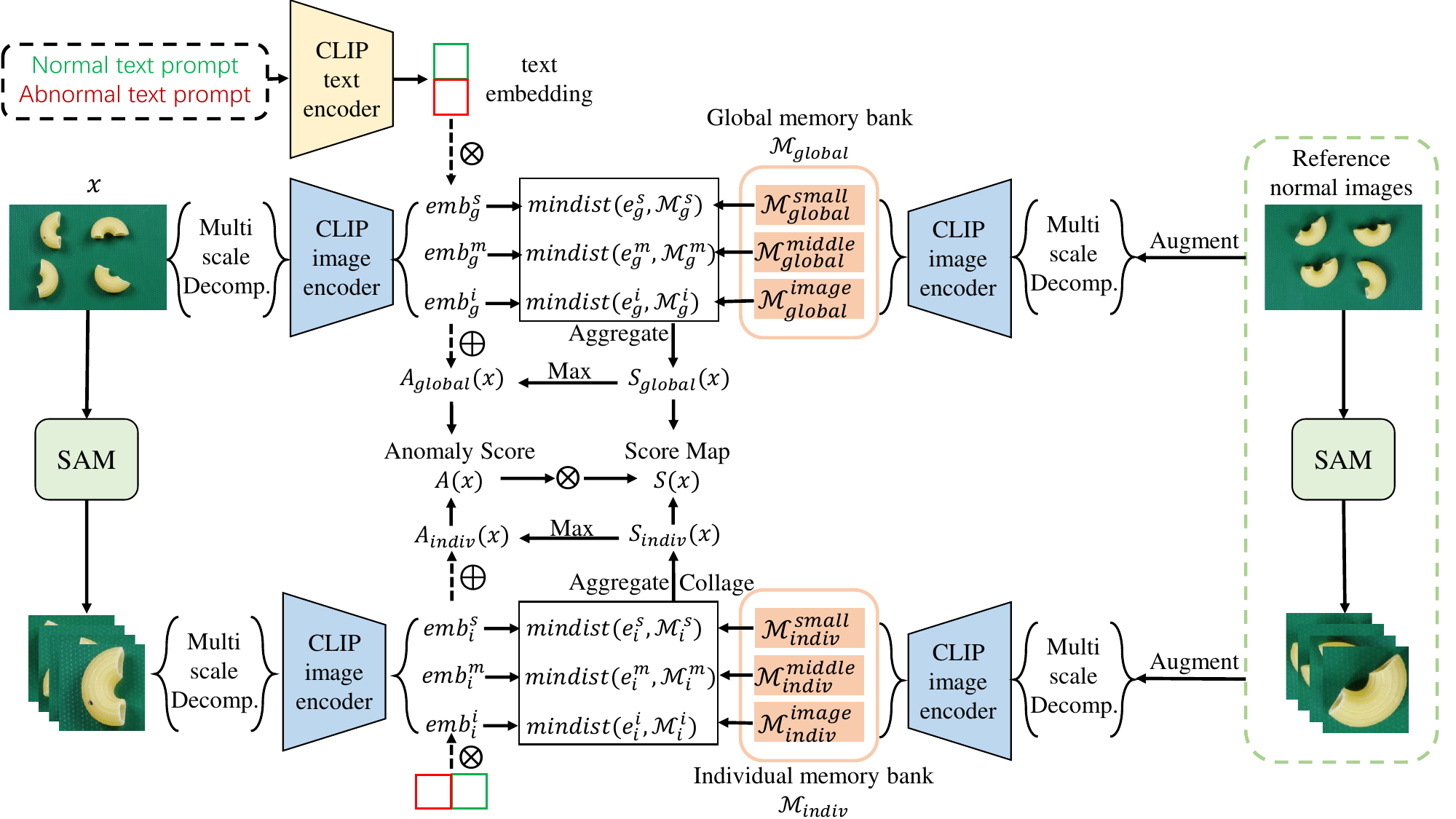}
    \caption{The overview of the proposed few-shot anomaly detection and localization architecture.}
    \label{fig:few}
\end{figure*}

\textbf{Memory Banks Construction:} We propose a memory-based framework for few-shot anomaly detection, as depicted in Figure~\ref{fig:few}. The framework incorporates two memory banks, constructed from global normal images and segmented individual images, to serve as references for segmenting anomalies in test images. The process begins by considering $k$ normal images denoted as $\{x_i\vert i = 1, ..., k\}$. Similar to the zero-shot scenario, individual images $\{x_{i,j}\}$ are segmented using SAM. To enhance the capacity of the memory bank, several augmentations, including flipping, rotation, and translation, are applied to the reference images before multi-scale decomposition.

To account for dissimilarities between test images and segmented images, we design separate global ($\mathcal{M}_{global}$) and individual ($\mathcal{M}_{indiv}$) memory banks. Each memory bank comprises three independent components (embeddings) obtained from multi-scale decomposition, including small $\mathcal{M}^{small}$, middle $\mathcal{M}^{middle}$, and image scales $\mathcal{M}^{image}$. Middle and small scale embeddings are concatenated by class tokens from different windows while image scale embeddings are patch-level output of CLIP image encoder.

\textbf{Memory Banks Comparison:} During the testing phase, the test image undergoes the same procedure as the reference images, but without any augmentation. Multi-scale embeddings generated from the test image are compared with the corresponding memory bank in cosine similarity. To retrieve the most similar embedding, a traversal process is employed, minimizing the distance $dist(e,\mathcal{M})$ between test embedding $e$ and reference memory bank $\mathcal{M}$ in the formula of
\begin{equation}
    dist(e, \mathcal{M}) =\min_{e_i\in \mathcal{M}} \frac{1-\cos{(e, e_i)}}{2},
\end{equation}
where,
\begin{equation}
\cos{(a,b)} = \frac{a\cdot b}{||a||||b||}.
\end{equation}

The anomaly score for a specific patch or window input is determined by evaluating the minimum distance between the test embedding and the memory embedding. For both the global and individual memory banks, the distances in small (s), middle (m) and image (i) scales are computed from the (test embedding, memory) pairs, namely $(e^s, \mathcal{M}^s), (e^m, \mathcal{M}^m), (e^i, \mathcal{M}^i)$, separately.

All pixels within the masked window are assigned the same anomaly score. The value of overlap area is calculated by the same harmonic averaging process as zero-shot pipeline. These different patch or window score maps are aggregated to obtain multi-scale score maps, which are then combined to form the global score map $S_{global}(x)$ and individual score map $S_{indiv}(x)$:
\begin{equation}
\begin{split}
S_{global}(x) = \lambda_{g,1} S_{g}^{s}(x) + \lambda_{g,2} S_{g}^{m}(x) + \lambda_{g,3} S_{g}^{i}(x),\\
S_{indiv}(x) = \lambda_{i,1} S_{i}^{s}(x) + \lambda_{i,2} S_{i}^{m}(x) + \lambda_{i,3} S_{i}^{i}(x),  
\end{split}
\end{equation}
where the superscript $s,m,i$ represents three scales and the subscript $g,i$ represents global and individual score map. $\lambda$ are trade-off parameters weighting the multi-scale score maps, we adopt $\lambda_{g,1} = \lambda_{g,1}= \lambda_{g,1} = 1, \lambda_{i,1} = 1.5, \lambda_{i,2} = 0.5, \lambda_{i,3} = 6$ as the default. It is important to note that the individual score map for multi-scale inputs is created by overlaying individual images onto their respective positions in the original image.

\textbf{Anomaly Detection:} Additionally, we leverage text prompt information, similar to the zero-shot approach, to assist in anomaly detection. This involves computing the anomaly scores of single-object images ($A_{single}(x)$) and multiple-object images ($A_{multi}(x)$). For the anomaly detection task, the final anomaly score $A(x)$ is obtained by combining the global and individual scores, each of which is composed of memory comparison and text alignment:
\begin{equation}
A(x) = A_{global}(x) + A_{indiv}(x),  
\end{equation}
where,
\begin{equation}
\begin{split}
A_{global}(x) = A_{multi}(x) + max_{i,j}S_{global}(x_{i,j}),\\
A_{indiv}(x) = A_{single}(x) + max_{i,j}S_{indiv}(x_{i,j}).  
\end{split}
\end{equation}

\textbf{Anomaly Segmentation:} For the anomaly segmentation task, we rely solely on memory comparison between the original image and individual images to generate the final score map $S(x)$. The final score map for test image is the combination of global score map $S_{global}$ and individual score map $S_{indiv}$, weighted by anomaly score $A(x)$,
\begin{equation}
    S(x) = A(x)\cdot (\lambda_{g} S_{global}(x) + \lambda_{i} S_{indiv}(x)).
\end{equation}
Here we take $\lambda_g = 0.55, \lambda_i = 0.45$. This approach effectively combines the benefits of memory-based comparison, text alignment, and multi-scale analysis to detect and segment anomalies in few-shot scenarios.

\section{Experiments}
\subsection{Implementation Details}
In this study, we have adopted the SAM (Segment Anything Model) approach, utilizing the official implementation provided by Facebook Research, as indicated by the reference\footnote{https://github.com/facebookresearch/segment-anything}. The pre-trained model selected for our experiments is the default ViT-H SAM model, which has been made publicly available. Additionally, we have employed the CLIP (Contrastive Language-Image Pretraining) model, trained on the LAION-400M dataset using the ViT-B/16+ architecture, as implemented in OpenCLIP\footnote{https://github.com/mlfoundations/open\_clip}.

In our experimental setup, we have set the stride of the sliding window for generating multi-scale images to 4, allowing for capturing a diverse range of image representations. To augment the fewshot memory bank, we have employed several concrete augmentation methods, including small-degree rotation, small-scale translation, horizontal flipping, and vertical flipping. Prior to inputting the images into the CLIP image encoder, we have resized all of them to a consistent dimension of $240\times 240$. Furthermore, it is important to note that all of our experiments were conducted on a single NVIDIA GTX 3090.

\subsection{Qualitative Results}

\begin{figure}
    \centering
    \includegraphics[width=0.48\textwidth]{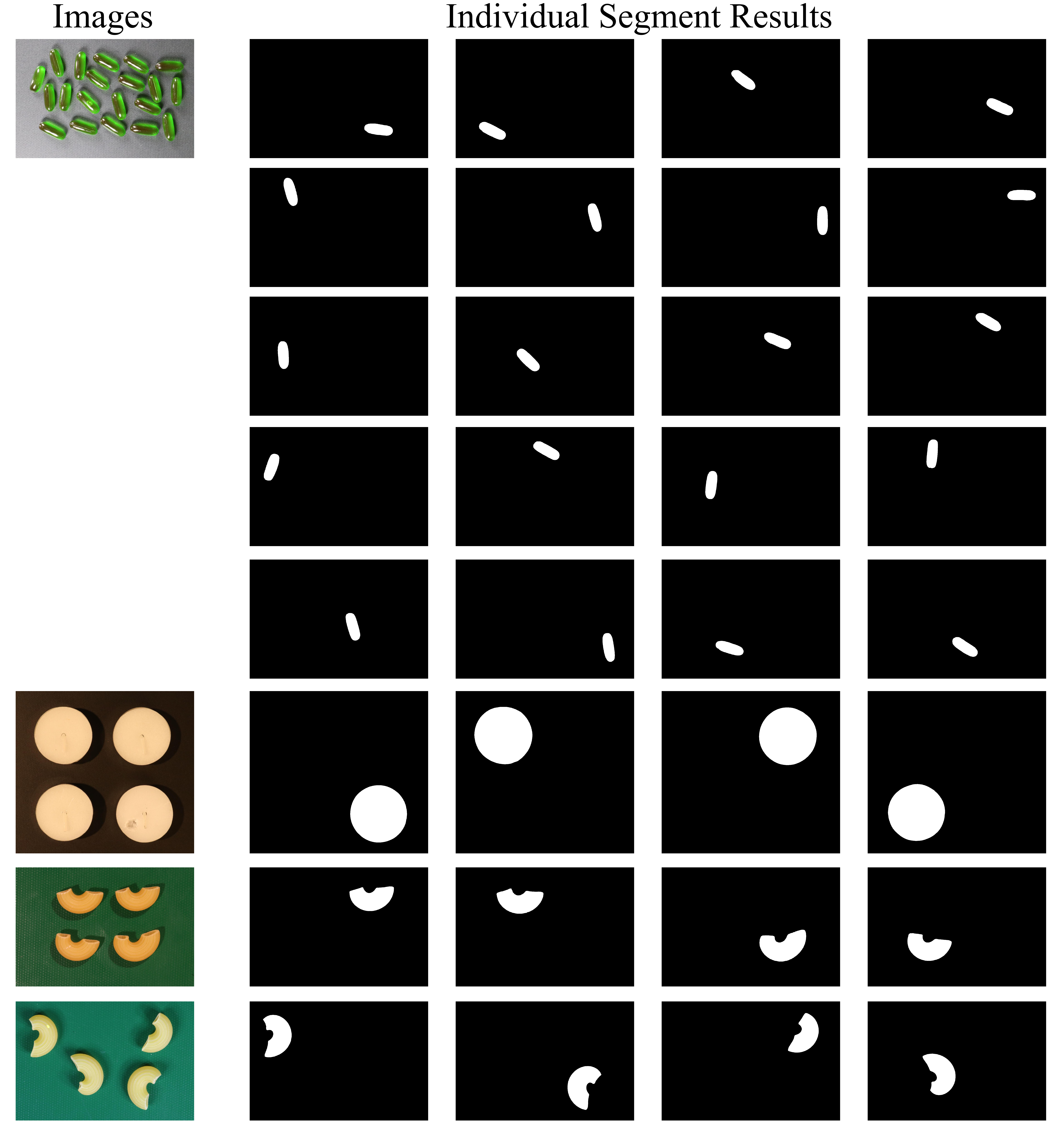}
    \caption{Visualization results of individual object segmentation in the VAND challenge.}
    \label{fig:sam_results}
\end{figure}

As shown in Figure~\ref{fig:sam_results}, we visualize the results of SAM for the individual object segmentation in the VAND challenge. We show results from images containing multiple objects, from the categories including capsules, candle, and macaroni. Results show that the individual objects are well separated in each case.

\subsection{Discussions}

The capacity of the memory bank plays a crucial role in determining the performance of our algorithm. A larger memory bank provides the advantage of accommodating a greater number of reference embeddings, allowing for a more thorough comparison between the test embedding and the reference embeddings. This expanded capacity facilitates a more comprehensive analysis and makes it easier to identify anomaly regions within the data.

However, when dealing with original images containing multiple objects, the segmentation process results in a larger number of segmented individual images. Consequently, the number of reference embeddings increases significantly, leading to higher memory occupancy of the memory bank. This heightened memory occupancy requires sufficient storage resources to accommodate the augmented volume of reference embeddings.

To address the constraints of GPU hardware and save computational time, it becomes necessary to control the capacity of the memory bank. When the number of given normal reference images is small (one to four), the memory bank constructed from all augmented images can fit within a single NVIDIA GTX 3090. However, with a greater number of few-shot images, especially in the case of multiple object images like candles and capsules, the memory capacity exceeds the capabilities of the device, making the execution time unmanageable.

To overcome this challenge, we adopt a solution where we randomly select a subset of reference embeddings to construct the memory bank. Specifically, regardless of the number of given normal images, we restrict the number of embeddings to 100,000. While this introduces an element of randomness and potentially lowers the overall algorithm performance, it effectively reduces the memory occupancy, allowing the algorithm to be executed within the limitations of available hardware resources.

Looking ahead, there are other potential methods worth exploring to further enhance the construction of the memory bank. For example, reducing the dimensionality of the reference embeddings or employing clustering algorithms to identify and record key embeddings could be valuable approaches. These alternatives may improve the algorithm's performance while optimizing memory utilization and computational efficiency.

\section{Conclusion}
In this paper, the multi-scale memory comparison framework offers a promising solution for zero-/few-shot anomaly detection in industrial contexts. By incorporating SAM segmentation, language prompts, and memory banks, our approach effectively handles complex scenes and delivers improved anomaly detection performance. The experimental results obtained further reinforce the effectiveness of our proposed method, highlighting its significance in the field of anomaly detection.


{\small
\bibliographystyle{ieee_fullname}
\bibliography{egbib}
}

\end{document}